\documentclass[runningheads]{llncs}
    \usepackage{siunitx}
    \usepackage{booktabs}
    \usepackage{amsmath}
    \usepackage{graphicx}
    \usepackage{subfig}
    \usepackage{tikz}
    \usetikzlibrary{positioning,backgrounds,arrows.meta}
    
\begin{document}
    \title{Recurrent DNNs and its Ensembles on the TIMIT Phone Recognition Task
    \thanks{Supported by Ministry of Education, Youth and Sports of the Czech Republic project No. LO1506 and by the grant of the University of West Bohemia, project No. SGS-2016-039. Access to computing and storage facilities owned by parties and projects contributing to the National Grid Infrastructure MetaCentrum provided under the programme "Projects of Large Research, Development, and Innovations Infrastructures" (CESNET LM2015042), is greatly appreciated.}}
    \author{Jan Van\v{e}k\orcidID{0000-0002-2639-6731} \and
    Josef Mich\'{a}lek\orcidID{0000-0001-7757-3163} \and \\
    Josef Psutka\orcidID{0000-0002-0764-3207}}
    \authorrunning{J. Van\v{e}k, J. Mich\'{a}lek, J. Psutka}
    \institute{University of West Bohemia \\ Univerzitn\'{i} 8, 301 00 Pilsen, Czech Republic\\
    \email{\{vanekyj,orcus,psutka\}@kky.zcu.cz}}
    \maketitle              % typeset the header of the contribution
    \begin{abstract}
    In this paper, we have investigated recurrent deep neural networks (DNNs) in combination with regularization techniques as dropout, zoneout, and regularization post-layer. As a benchmark, we chose the TIMIT phone recognition task due to its popularity and broad availability in the community. It also simulates a low-resource scenario that is helpful in minor languages. Also, we prefer the phone recognition task because it is much more sensitive to an acoustic model quality than a large vocabulary continuous speech recognition task. In recent years, recurrent DNNs pushed the error rates in automatic speech recognition down. But, there was no clear winner in proposed architectures. The dropout was used as the regularization technique in most cases, but combination with other regularization techniques together with model ensembles was omitted. However, just an ensemble of recurrent DNNs performed best and achieved an average phone error rate from 10 experiments 14.84~\% (minimum 14.69~\%) on core test set that is slightly lower then the best-published PER to date, according to our knowledge. Finally, in contrast of the most papers, we published the open-source scripts to easily replicate the results and to help continue the development. 
    
    \end{abstract}
    \section{Introduction}
    Phone recognition on the Texas Instruments/Massachusetts Institute of Technology (TIMIT) corpus of read speech \cite{TIMIT} is very popular benchmark task. The phone recognition is much more sensitive to quality of the acoustic model than a large vocabulary continuous speech recognition (LVCSR). Therefore it is a good benchmark to test novel DNN architectures and training strategies. The TIMIT corpus has defined training, development, and two test sets. It helped to achieve results comparability over publications.
    
    The first generation of DNNs was based on feed forward networks. Then recurrent DNNs took the lead in the last few years. First, Mohamed at al. presented its monophone deep belief network (DBN) \cite{Mohammed_TASL_2011} with PER 20.7\% on the core test set. A triphone version of the DBN with a speaker adaptive training and a fMLLR adaptation was developed by Bagher BabaAli and Karel Vesely in the TIMIT Kaldi example s5 \cite{Kaldi_GIT}. The Kaldi example achieved PER 18.5\% on the core test set. Better results were then obtained by DNNs with rectified linear units (ReLU). The ReLU DNNs do not need the DBN pretraining and, if dropout is applied, they perform well on held out data. Laszlo Toth reported a PER 17.76\% on the core test set with a convolutional bottle neck ReLU DNN in \cite{Toth_IS2013} and a year later he reported PER 16.5\% with a 2D convolutional bottleneck maxout DNN in \cite{Toth_IS2014}.
    %zmenit na "We" ve finalni verzi
    Vanek also reported PER 16.5\% with an ensemble of DBN DNNs augmented by regularization post-layer \cite{Vanek_2017}. Taesup Moon then achieved stable PER 16.9\% with a dropout bi-directional long-short term memory recurrent DNN (DBLSTM) and a peak PER with a larger net up to 16.29\% \cite{Moon2015}. Ravanelli at al. presented the best actual result, to our knowledge, PER 14.9\% with a bi-directional modified gated recurrent unit (GRU) based DNN \cite{Ravanelli_IS2017}.
    
    In this paper, we have investigated recurrent deep neural networks in combination with regularization techniques as dropout, zoneout, and regularization post-layer (RPL). For comparison, we have evaluated a simple feed-forward ReLU DNN also. Moreover, we published our scripts to easily repeat our work and results. We followed the Kaldi s5 example and limit the experiments to a triphone model obtained by the Kaldi example together with the fMLLR speaker adapted training, development, and test data. Then, we have trained various DNNs. Because of the common feature processing stage, we did not try any 2D convolutional DNNs. We plan to investigate them as a future work.

    \section{Recurrent Neural Network Architectures}
    \subsection{Long Short-Term Memory}
\label{sec:lstm}
Long short-term memory (LSTM) is a widely used type of recurrent neural network (RNN).
Standard RNNs suffer from both exploding and vanishing gradient problems.
%Both of these problems are caused by the fact, that information flowing through the RNN passes through many stages of multiplication.
%The gradient is essentially equal to the weight matrix raised to a high power.
%This results in the gradient growing or shrinking at an exponential rate to the number of timesteps.

The exploding gradient problem can be solved simply by truncating the gradient.
On the other hand, the vanishing gradient problem is harder to overcome.
It does not simply cause the gradient to be small; the gradient components corresponding to long-term dependencies are small while the components corresponding to short-term dependencies are large.
%Resulting RNN can then learn short-term dependencies but not long-term dependencies.

The LSTM was proposed in 1997 by Hochreiter and Scmidhuber \cite{hochreiter1997long} as a solution to the vanishing gradient problem.
Let $c_t$ denote a hidden state of a LSTM.
The main idea is that instead of computing $c_t$ directly from $c_{t-1}$ with matrix-vector product followed by an activation function, the LSTM computes $\Delta c_t$ and adds it to $c_{t-1}$ to get $c_t$.
The addition operation is what eliminates the vanishing gradient problem.

Each LSTM cell is composed of smaller units called gates, which control the flow of information through the cell.
The forget gate $f_t$ controls what information will be discarded from the cell state, input gate $i_t$ controls what new information will be stored in the cell state and output gate $o_t$ controls what information from the cell state will be used in the output.

The LSTM has two hidden states, $c_t$ and $h_t$.
The state $c_t$ fights the gradient vanishing problem while $h_t$ allows the network to make complex decisions over short periods of time.
There are several slightly different LSTM variants.
The architecture used in this paper is specified by the following equations:
\begin{align*}
    i_t &= \sigma(W_{xi} x_t + W_{hi} h_{t-1} + b_i) \\
    f_t &= \sigma(W_{xf} x_t + W_{hf} h_{t-1} + b_f) \\
    o_t &= \sigma(W_{xo} x_t + W_{ho} h_{t-1} + b_o) \\
    c_t &= f_t \ast c_{t-1} + i_t \ast \tanh(W_{xc} x_t + W_{hc} h_{t-1} + b_c) \\
    h_t &= o_t \ast \tanh(c_t)
\end{align*}

\subsection{Gated Recurrent Unit}
A gated recurrent unit (GRU) was proposed in 2014 by Cho et al.\cite{chung2014empirical}
Similarly to the LSTM unit, the GRU has gating units that modulate the flow of information inside the unit, however, without having separate memory cells.

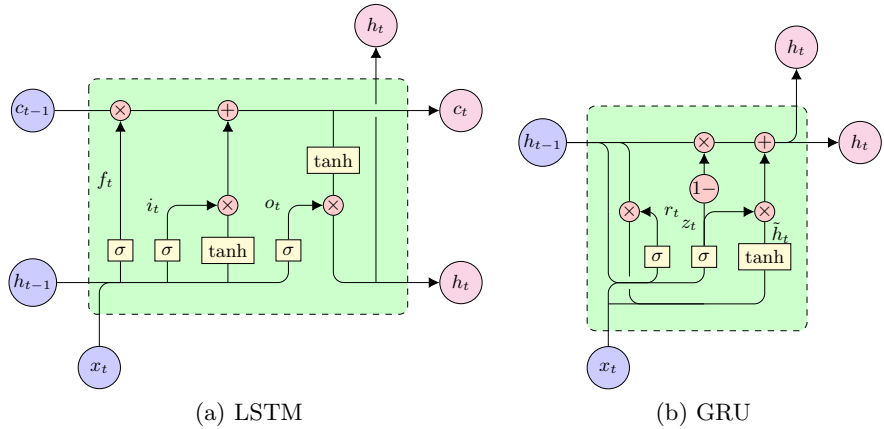
\begin{figure}
    \centering
    \subfloat[LSTM]{
    \scalebox{0.8}{
        \begin{tikzpicture}[node distance=10pt]
    \tikzstyle{phantom}=[inner sep=0,outer sep=0,minimum size=0]
    \tikzstyle{pointwise}=[draw,circle,fill=red!20!white,inner sep=0]
    \tikzstyle{layer}=[draw,rectangle,fill=yellow!20!white]
    \tikzstyle{transition}=[-{Latex[length=5pt,width=5pt]},rounded corners=5]
    \tikzstyle{line}=[rounded corners=5]
    \tikzstyle{node_in}=[draw,circle,fill=blue!20!white,inner sep=1pt,minimum size=20pt]
    \tikzstyle{node_out}=[draw,circle,fill=magenta!20!white,inner sep=1pt,minimum size=20pt]

    % main nodes
    \node[layer] (forget_gate) {$\sigma$};
    \node[layer,right=of forget_gate] (input_gate) {$\sigma$};
    \node[layer,right=of input_gate] (input_tanh) {$\tanh$};
    \node[layer,right=of input_tanh] (output_gate) {$\sigma$};
    \node[pointwise,above=of input_tanh] (input_mult) {$\times$};
    \node[pointwise,above=35pt of input_mult] (input_add) {$+$};
    \node[pointwise,right=40pt of input_mult] (output_mult) {$\times$};
    \node[layer,above=of output_mult] (output_tanh) {$\tanh$};

    % nodes below
    \node[phantom,below=of forget_gate] (forget_below) {};
    \node[phantom] at(forget_below -| input_gate) (input_below) {};
    \node[phantom] at(forget_below -| input_tanh) (input_tanh_below) {};
    \node[node_in,left=30pt of forget_below] (h_in) {$h_{t-1}$};
    \node[phantom,right=70pt of input_tanh_below] (output_below) {};
    \node[node_out,right=30pt of output_below] (h_out) {$h_t$};
    \node[node_in,below=30pt of forget_below,xshift=-10pt] (input_x) {$x_t$};

    % nodes above
    \node[pointwise] at (forget_gate |- input_add) (forget_mult) {$\times$};
    \node[node_in] at(forget_mult -| h_in) (c_in) {$c_{t-1}$};
    \node[phantom] at(forget_mult -| output_tanh) (output_tanh_above) {};
    \node[node_out] at(forget_mult -| h_out) (c_out) {$c_t$};
    \node[phantom] at(forget_mult -| output_below) (output_above) {};
    \node[node_out,above=30pt of output_above] (output_h) {$h_t$};

    % main transitions
    \draw[transition] (forget_gate) -- (forget_mult) node[midway,left] {$f_t$};
    \draw[transition] (input_gate) |- (input_mult) node[midway,left] {$i_t$};
    \draw[line] (input_tanh) -- (input_mult);
    \draw[transition] (input_mult) -- (input_add);
    \draw[transition] (output_gate) |- (output_mult) node[midway,left] {$o_t$};
    \draw[line] (output_tanh_above) -- (output_tanh);
    \draw[line] (output_tanh) -- (output_mult);

    % transitions below
    \draw[line] (h_in) -| (output_gate);
    \draw[line] (forget_below) -- (forget_gate);
    \draw[line] (input_below) -- (input_gate);
    \draw[line] (input_tanh_below) -- (input_tanh);
    \draw[transition] (output_mult) |- (h_out);
    \draw[line] (input_x) |- (forget_below);

    % transitions above
    \draw[line] (c_in) -- (forget_mult);
    \draw[line] (forget_mult) -- (input_add);
    \draw[transition] (input_add) -- (c_out);
    \draw[line,shorten >=3pt] (output_below) -- (output_above);
    \draw[transition,shorten <=3pt] (output_above) -- (output_h);

    \begin{scope}[on background layer]
        \path (forget_mult)+(-15pt,15pt) node (a) {};
        \path (output_below)+(15pt,-15pt) node (b) {};
        \path[draw,dashed,fill=green!20!white,rounded corners] (a) rectangle (b);
    \end{scope}
\end{tikzpicture}
        }
    }
    \subfloat[GRU]{
    \scalebox{0.8}{
        \begin{tikzpicture}[node distance=10pt]
    \tikzstyle{phantom}=[inner sep=0,outer sep=0,minimum size=0]
    \tikzstyle{pointwise}=[draw,circle,fill=red!20!white,inner sep=0]
    \tikzstyle{layer}=[draw,rectangle,fill=yellow!20!white]
    \tikzstyle{transition}=[-{Latex[length=5pt,width=5pt]},rounded corners=5]
    \tikzstyle{line}=[rounded corners=5]
    \tikzstyle{node_in}=[draw,circle,fill=blue!20!white,inner sep=1pt,minimum size=20pt]
    \tikzstyle{node_out}=[draw,circle,fill=magenta!20!white,inner sep=1pt,minimum size=20pt]

    % main nodes
    \node[layer] (reset_gate) {$\sigma$};
    \node[layer,right=of reset_gate] (update_gate) {$\sigma$};
    \node[layer,right=of update_gate] (tanh) {$\tanh$};
    \node[pointwise,above=of tanh] (tanh_mult) {$\times$};
    \node[pointwise,above=20pt of update_gate] (minus_one) {$1-$};
    \node[pointwise,above=of minus_one] (minus_mult) {$\times$};
    \node[pointwise] at(minus_mult -| tanh_mult) (tanh_add) {$+$};

    % nodes below
    \node[phantom,below left=of reset_gate] (reset_below) {};
    \node[phantom,left=of reset_below] (reset_belowl) {};
    \node[phantom,below=of reset_below] (reset_belowd) {};
    \node[phantom] at(reset_belowl |- reset_belowd) (reset_belowld) {};
    \node[phantom] at(update_gate |- reset_belowd) (reset_belowrd) {};

    % main nodes
    \node[pointwise] at(tanh_mult -| reset_below) (reset_mult) {$\times$};
    \node[node_in,below=20pt of reset_belowld] (input_x) {$x_t$};
    \node[node_out,right=30pt of tanh_add] (h_out) {$h_t$};
    \node[node_out,above=30pt of tanh_add,xshift=15pt] (output_h) {$h_t$};

    % nodes above
    \node[phantom] at(reset_mult -| reset_belowl) (reset_above) {};
    \node[node_in,left=60pt of minus_mult] (h_in) {$h_{t-1}$};

    % main transitions
    \draw[transition] (reset_gate) |- (reset_mult) node[midway,right] {$r_t$};
    \draw[line] (update_gate) -- (minus_one) node[midway,left] {$z_t$};
    \draw[line] (tanh) -- (tanh_mult) node[midway,right] {$\tilde{h}_t$};
    \draw[transition] (minus_one) -- (minus_mult);
    \draw[transition] (tanh_mult) -- (tanh_add);
    \draw[transition] (update_gate) |- (tanh_mult);

    % transitions below
    \draw[line] (reset_below) -| (reset_gate);
    \draw[line] (reset_below) -| (update_gate);
    \draw[line] (input_x) |- (reset_below);
    \draw[line] (reset_belowld) -| (tanh);
    \draw[line] (reset_above) |- (reset_below);
    \draw[line,shorten >=3pt] (reset_mult) -- (reset_below);
    \draw[line,shorten >=3pt] (reset_belowrd) -| (reset_below);

    % transitions above
    \draw[line] (h_in) -| (reset_above);
    \draw[line] (h_in) -| (reset_mult);
    \draw[line] (h_in) -- (minus_mult);
    \draw[line] (minus_mult) -- (tanh_add);
    \draw[transition] (tanh_add) -- (h_out);
    \draw[transition] (tanh_add) -| (output_h);

    \begin{scope}[on background layer]
        \path (reset_above)+(-10pt,50pt) node (a) {};
        \path (tanh)+(20pt,-35pt) node (b) {};
        \path[draw,dashed,fill=green!20!white,rounded corners] (a) rectangle (b);
    \end{scope}
\end{tikzpicture}
        }
    }
    \caption{Structure of LSTM and GRU units\cite{Olah2015}}
    \label{fig:lstmgru}
\end{figure}

The update gate $z_t$ decides how much the unit updates its activation and reset gate $r_t$ determines which information will be kept from the old state.
GRU does not have any mechanism to control what information to output, therefore it exposes the whole state.
The figure \ref{fig:lstmgru} shows the internal structure of LSTM and GRU units.

The main differences between LSTM unit and GRU are:
\begin{itemize}
    \item GRU has 2 gates, LSTM has 3 gates
    \item GRUs do not have an internal memory different from the unit output, LSTMs have an internal memory $c_t$ and the output is controlled by an output gate
    \item Second nonlinearity is not applied when computing the output of GRUs
\end{itemize}

The GRU unit used in this work is described by the following equations:
\begin{align*}
    r_t &= \sigma(W_r x_t + U_r h_{t - 1} + b_r) \\
    z_t &= \sigma(W_z x_t + U_z h_{t - 1} + b_z) \\
    \tilde{h_t} &= \tanh(W x_t + U (r_t \ast h_{t - 1}) + b_h) \\
    h_t &= (1 - z_t) \ast h_{t-1} + z_t \ast \tilde{h_t} \\
\end{align*}

    \section{Regularization Techniques}
    It is well know that proper regularization techniques are needed in deep neural network systems. DNNs work very well on training data but performance on held out data may be much worse. A proper combination of dropout, early stopping, and L2 regularization has developed as a standard in feed forward DNNs. Using of dropout in recurrent DNNs may be more tricky. An interesting analysis focused on the acoustic modeling domain was published in \cite{Cheng_2017}. An alternative technique called zoneout was published in \cite{Krueger_2017}. The per-network regularization techniques may also be combined by using network ensembles. The techniques used in this paper are described below, in more detail.
 
\subsection{Dropout}
Dropout \cite{srivastava2014dropout} consists of multiplying neural net activations by random zero-one masks during training. The dropout probability p determines what proportion of the mask values are one. Usually, the dropout rate p = 0.5 and it works well for a range of tasks. But it is not an ideal setup for acoustic modeling. In contrast, for example, p = 0.2 was suggested in \cite{Vanek_2017} and \cite{Toth_IS2013} for TIMIT phone recognition task. In \cite{Cheng_2017}, a per epoch dynamic schedule was proposed. They used a constant number of epochs. The training started with 20\% of the epochs having p = 0. Then, there was a peek of p = 0.1 to 0.15 in the middle of epochs and the training ended again with p = 0. p-values for the intermediate epochs are linearly interpolated.

\subsection{Zoneout}
The alternative technique for recurrent DNNs was published in \cite{Krueger_2017}. At each timestep, zoneout stochastically forces some hidden units to maintain their previous values. Like dropout, zoneout uses random noise to train a pseudo-ensemble, improving generalization. But by preserving instead of dropping hidden units, gradient information and state information are more readily propagated through time, as in feed forward stochastic depth networks. However, the technique was tested on a permuted sequential MNIST image recognition task not the speech or phone recognition. 

\subsection{Cross-Validation Aggregation - Crogging}
In classification, cross-validation (CV) is widely employed to estimate the expected accuracy of a predictive algorithm by averaging predictive errors across mutually exclusive sub-samples of the data. Similarly, bootstrapping (Bagging) aims to increase the validity of estimating the expected accuracy by repeatedly sub-sampling the data with replacement, creating overlapping samples of the data \cite{breiman1996bagging}. Beyond error estimation, bootstrap aggregation or bagging is used to make a NNs ensemble. Barrow et al. considered in \cite{Barrow2013} similar extensions of cross-validation to create diverse models. By bagging, it was proposed to combine the benefits of cross-validation and prediction aggregation, called crogging. In \cite{Barrow2013}, the crogging approach significantly improved prediction accuracy relative to bagging.

\subsection{Regularization Post-Layer (RPL)}
\label{RPL}
The RPL technique is based on cross-validation aggregation \cite{Vanek_2017}; however, it is more advanced. Cross-validation ensures that all observations are used for both training and validation, though not simultaneously, and each observation is guaranteed to be used for model estimation and validation the same number of times. Furthermore, the validation set available in CV can be used to control for overfitting in neural network training using early stopping. A $k$-fold cross-validation allows the use of all $k$ validation sets in performing early stopping, and this potentially further reduces the risk of overfitting. Moreover, prediction values for all $k$ folds can be obtained in validation mode. These validation-predictions for all folds -- all training data -- make a new valuable training set for an additional NN layer that is called regularization post-layer. The RPL input dimension is equal to the number of classes, and the output dimension is the same. The RPL uses log-softmax values of the main network softmax output and ends also with softmax. Because of large number of classes in acoustic models, the RPL with diagonal matrix only is preferred.

    \section{Experiments}
    
    The TIMIT corpus contains recordings of phonetically-balanced prompted English speech. It was recorded using a Sennheiser close-talking microphone at 16 kHz rate with 16 bit sample resolution. TIMIT contains a total of 6300 sentences (5.4 hours), consisting of 10 sentences spoken by each of 630 speakers from 8 major dialect regions of the United States. All sentences were manually segmented at the phone level.
    
    The prompts for the 6300 utterances consist of 2 dialect sentences (SA), 450 phonetically compact sentences (SX) and 1890 phonetically-diverse sentences (SI).
    
    The training set contains 3696 utterances from 462 speakers. The core test set consists of 192 utterances, 8 from each of 24 speakers (2 males and 1 female from each dialect region). The training and test sets do not overlap. 
    \subsection{Speech Data, Processing, and Test Description}
    As mentioned above, we used TIMIT data available from LDC as a corpus LDC93S1. Then, we ran the Kaldi TIMIT example script s5, which trained various NN-based phone recognition systems with a common HMM-GMM tied-triphone model and alignments. The common baseline system consisted of the following methods: It started from MFCC features which were augmented by $\Delta$ and $\Delta\Delta$ coefficients and then processed by LDA. Final feature vector dimension was 40. We obtained final alignments by HMM-GMM tied-triphone model with 1909 tied-states (may vary slightly if the script is re-run). We trained the model with MLLT and SAT methods, and we used fMLLR for the SAT training and a test phase adaptation. We dumped all training, development and test fMLLR processed data, and alignments to disk. Therefore, it was easy to do compatible experiments from the same common starting point. We employed a bigram language/phone model for final phone recognition. A bigram model is a very weak model for phone recognition; however, it forced focus to the acoustic part of the system, and it boosted benchmark sensitivity. The training, as well as the recognition, was done for 48 phones. We mapped the final results on TIMIT core test set to 39 phones (as it is usual by processing TIMIT corpus), and phone error rate (PER) was evaluated by the provided NIST script to be compatible with previously published works. In contrast to the Kaldi recipe, we used a different phone decoder. It is a standard Viterbi-based triphone decoder. It gives better results than the Kaldi standard WFST decoder on the TIMIT phone recognition task.
    We have used an open-source Chainer 3.0 DNNs Python tranining tool that supports NVidia GPUs \cite{Chainer}. It is multiplatform and easy to use. 
    
    %\subsection{Feed-Forward DNNs}
    
    %\subsection{Recurrent DNNs}
    \subsection{DNN Training and Results}
    First as a reference to RNNs, we trained feed-forward (FF) DNN with ReLU activation function without any pre-training. We used dropout $p=0.2$. We stacked 11 input fMLLR feature frames to 440 NN input dimension, like in Kaldi example s5. All the input vectors were transformed by an affine transform to normalize input distribution. We have used a network with 8 hidden layers and 2048 ReLU neurons, because it gave the best performance according to our preliminary experiments. The final softmax layer had 1909 neurons. We used SGD with momentum 0.9. The learning rate was three-times reduced according to development data training criterion change. Together with the learning rate reduction, the batch size was gradually increased from initial 256 to 1024, and final 2048.
    
    \begin{table}
        \center
        \caption{DNN Phone Error Rate [\%]}
        \label{tbl:per}
        \setlength{\tabcolsep}{1.5mm}
        \begin{tabular}{l|ccccc}
            \toprule
             & FF & LSTM & GRU & Zoneout LSTM\\
            \midrule
            Master & $17.00 \pm 0.23$ & $15.30 \pm 0.13$ & $15.66 \pm 0.19$ & $21.73 \pm 0.26$\\
            Master + RPL & $17.09 \pm 0.26$ & $15.29 \pm 0.21$ & $15.71 \pm 0.14$ & $27.81 \pm 0.40$\\
            Folds & $17.14 \pm 0.09$ & $14.98 \pm 0.10$ & $15.12 \pm 0.13$ & $20.98 \pm 0.19$\\
            Folds + RPL & $17.27 \pm 0.10$ & $14.94 \pm 0.12$ & $15.27 \pm 0.13$ & $28.73 \pm 0.20$\\
            Master + Folds & $17.04 \pm 0.10$ & $14.84 \pm 0.14$ & $15.22 \pm 0.11$ & $20.81 \pm 0.19$\\
            Master + Folds + RPL & $17.17 \pm 0.09$ & $14.84 \pm 0.12$ & $15.22 \pm 0.09$ & $28.17 \pm 0.24$\\
            \bottomrule
        \end{tabular}
    \end{table}

    Then we have trained LSTM, GRU and Zoneout LSTM networks. For all of these recurrent networks, we have used identical training setup. The dropout used was $p=0.2$. We have used output time delay equal to 5 time steps. RNNs were trained in 4 stages. The first stage used Adam optimization algorithm with batch size 512. The other stages used SGD with momentum 0.9, batch size 128 and learning rate equal to \num{1e-3}, \num{1e-4}, and \num{1e-5} respectivelly. The training in each stage was stopped when the development data criterion increased in comparison to the last epoch.
    In zoneout LSTM case, we used parameters with values $d^c = d^h = 0.5$.
    
    We have trained each network in several scenarios.
    First, we have trained a single network from the whole training set called \emph{Master} network.
    Then, we have used the crogging technique: we have divided the training set into 5 folds and trained 5 networks for each set of all folds except one.
    In the evaluation phase, we have used the average output of the ensemble of the 5 fold networks called \emph{Folds} in the result table.
    Also, we have evaluated the combination of the \emph{Master} and \emph{Folds} network ensemble, with \emph{Master} having weight 50~\%.
    Finally, we have then trained the RPL layer for the fold network ensemble and evaluated it with all 3 variants (only \emph{Master}, only \emph{Folds}, and \emph{Master} with \emph{Folds}).
    
    Each experiment was performed 10 times in total and the phone error rate (PER) was then evaluated.
    Table \ref{tbl:per} shows the average phone error rate and standard deviation for each experiment.
    From the table, it is obvious that the lowest PER in all scenarios was obtained with LSTM.
    GRU gave the second best results, in average higher by 0.33~\%.
    We have received the worst results with zoneout LSTM.
    They are significantly worse than all other networks and we were not able to improve them.
    In all RNNs used, \emph{Folds} network ensemble gave better performance than using only single \emph{Master} network.
    Also, the combination of the \emph{Master} network and \emph{Folds} network ensemble further improved the performance in both LSTM and zoneout LSTM cases.
    Slight improvements were also obtained in a few cases by employing RPL, most notably almost all LSTM scenarios and some GRU experiments with \emph{Master} and \emph{Folds}.
    Although the best average PER from all experiments we have obtained is 14.84~\%, the best single experiment was LSTM with \emph{Master}, \emph{Folds} and RPL, which resulted in 14.64~\% PER.

    From all our experiments, we have found that using \emph{Folds} network ensemble generally leads to better performance in RNNs, while RPL can give better results only in some cases.
    Disadvantage of using \emph{Folds} network ensemble is that it is more computationally intensive, because several networks have to be evaluated, while RPL gives almost no overhead once trained.
    
    \section{Conclusion}
    We have trained several neural network architectures and evaluated their performance on the phone recognition task with or without model ensembles and regularization post-layer (RPL).
    We have evaluated feed-forward DNNs and RNNs composed of LSTM, GRU, or zoneout LSTM units.
    Our experiments showed, that model ensembles give better performance than using a single network in all RNN cases.
    Also, the combination of a single \emph{Master} network and model ensembles further improved performance in LSTM and zoneout LSTM networks.
    Some improvements can also be gained from using RPL, although it led to better phone error rate (PER) only in LSTM and some GRU scenarios.
    The best average PER we have obtained is 14.84~\% in LSTM network with \emph{Master} network and ensemble models, which is slightly lower than the best-published PER to date, according to our knowledge.
    The best single experiment resulted in 14.64~\% PER and it was the LSTM network with \emph{Master} network, ensemble models and RPL.

    We have used Chainer 3.0 DNN training framework with Python programming language and all our scripts used in this work are publicly available at \url{https://github.com/OrcusCZ/NNAcousticModeling}.

    \bibliographystyle{splncs04}
    \bibliography{SPECOM2018_Vanek_Michalek}
    \end{document}